\documentclass[sigconf]{acmart}
\AtBeginDocument{%
  }

\copyrightyear{2025}
\acmYear{2025}
\setcopyright{cc}
\setcctype{by}
\acmConference[SIGSPATIAL '25]{The 33rd ACM International Conference on Advances in Geographic Information Systems}{November 3--6, 2025}{Minneapolis, MN, USA}
\acmBooktitle{The 33rd ACM International Conference on Advances in Geographic Information Systems (SIGSPATIAL '25), November 3--6, 2025, Minneapolis, MN, USA}
\acmDOI{10.1145/3748636.3762716}
\acmISBN{979-8-4007-2086-4/2025/11}

\usepackage{amsmath}
\usepackage[capitalize]{cleveref}
\usepackage{amsfonts}
\usepackage{bm}
\usepackage{multirow}
\usepackage{booktabs}  
\usepackage{makecell}
\usepackage{xcolor}
\usepackage{natbib}
\usepackage{makecell} 
\usepackage{enumitem}
\usepackage[table]{xcolor}
\usepackage{float}  
\usepackage{balance}
\usepackage{stfloats}

\begin{document}

\title{Geo-Aware Models for Stream Temperature Prediction across Different Spatial Regions and Scales}

\author{Shiyuan Luo}
\affiliation{
  \institution{University of Pittsburgh}
  \country{}
}
\email{shl298@pitt.edu}

\author{Runlong Yu}
\affiliation{
  \institution{University of Alabama}
\country{}
}
\email{ryu5@ua.edu}

\author{Shengyu Chen}
\affiliation{%
  \institution{University of Pittsburgh}
  \country{}
}
\email{shc160@pitt.edu}

\author{Yingda Fan}
\affiliation{%
 \institution{University of Pittsburgh}
 \country{}
 }
\email{yif47@pitt.edu}

\author{Yiqun Xie}
\affiliation{%
  \institution{University of Maryland}
  \country{}
}
\email{xie@umd.edu}

\author{Yanhua Li}
\affiliation{%
  \institution{Worcester Polytechnic Institute}
  \country{}
  }
\email{yli15@wpi.edu}

\author{Xiaowei Jia}
\affiliation{%
  \institution{University of Pittsburgh}
  \country{}
}
\email{xiaowei@pitt.edu}

\renewcommand{\shortauthors}{Luo et al.}

\begin{abstract}
Understanding environmental ecosystems is vital for the sustainable management of our planet. However,
existing physics-based and data-driven models often fail to generalize to varying spatial regions and scales due to the inherent data heterogeneity presented in real environmental ecosystems. This generalization issue is further exacerbated by the limited observation samples available for model training. 
To address these issues, we propose Geo-STARS, a geo-aware spatio-temporal modeling framework for predicting stream water temperature across different watersheds and spatial scales. The major innovation of Geo-STARS is the introduction of geo-aware embedding, which leverages geographic information to explicitly capture shared principles and patterns across spatial regions and scales.  We further integrate the geo-aware embedding into a gated spatio-temporal graph neural network. This design enables the model to learn complex spatial and temporal patterns guided by geographic and hydrological context, even with sparse or no observational data. 
We evaluate Geo-STARS's efficacy in predicting stream water temperature, which is a master factor for water quality.  
Using real-world datasets spanning 37 years across multiple watersheds along the eastern coast of the United States, Geo-STARS demonstrates its superior generalization performance across both regions and scales, outperforming state-of-the-art baselines.
These results highlight the promise of Geo-STARS for scalable, data-efficient environmental monitoring and decision-making.
\end{abstract}




\maketitle

\section{Introduction}
Understanding the dynamics of environmental ecosystems is critical for sustainable management of natural resources and mitigating natural disasters such as algal blooms and floods~\cite{yu2025environmental}. 
Accurate prediction of key environmental variables and their dynamics can help resource managers make better decisions, which is especially important due to the increasing demands on food, water, and energy caused by population growth and changing climate 
~\cite {hall2014coping,mbow2020food,konar2011water}. 
For example, drinking water reservoir operators in the Delaware River Basin need to supply safe drinking water to New York City while also maintaining sufficient streamflow and cool water temperatures for fish habitat in the rivers downstream of the reservoirs~\cite{williamson2015summary}.

Given the importance of this problem, scientists have been developing physics-based models (also called process-based models) to simulate underlying physical processes for different components of ecosystems~\cite{hipsey_general_2019,markstrom2012p2s}. Despite their extensive use, most physics-based models are necessarily approximations of reality due to incomplete knowledge or excessive complexity in modeling certain processes~\cite{gupta2014debates}.
Substantial computational costs of these models pose a major impediment to simulating physical processes at fine spatial resolutions required for effective analysis in many real-world~\cite{karpatne2024knowledge}.
Moreover, these models often need to be recalibrated when deployed in different spatial regions~\cite{kratzert2019benchmarking_backup}. 
More recently, machine learning (ML) models have been increasingly considered as an alternative approach for environmental modeling, given their computational efficiency in the prediction phase and their ability to extract complex relationships from input drivers to the target variable~\cite{shen2021applications,chen2023meta}.  
Recent advances in spatio-temporal models have also shown promise in capturing complex spatial-temporal correlations of different environmental processes~\cite{sun2021explore,jia2021physics_sdm,chen2022physics}.

\begin{figure}[t]
\centering
\includegraphics[width=1\linewidth]{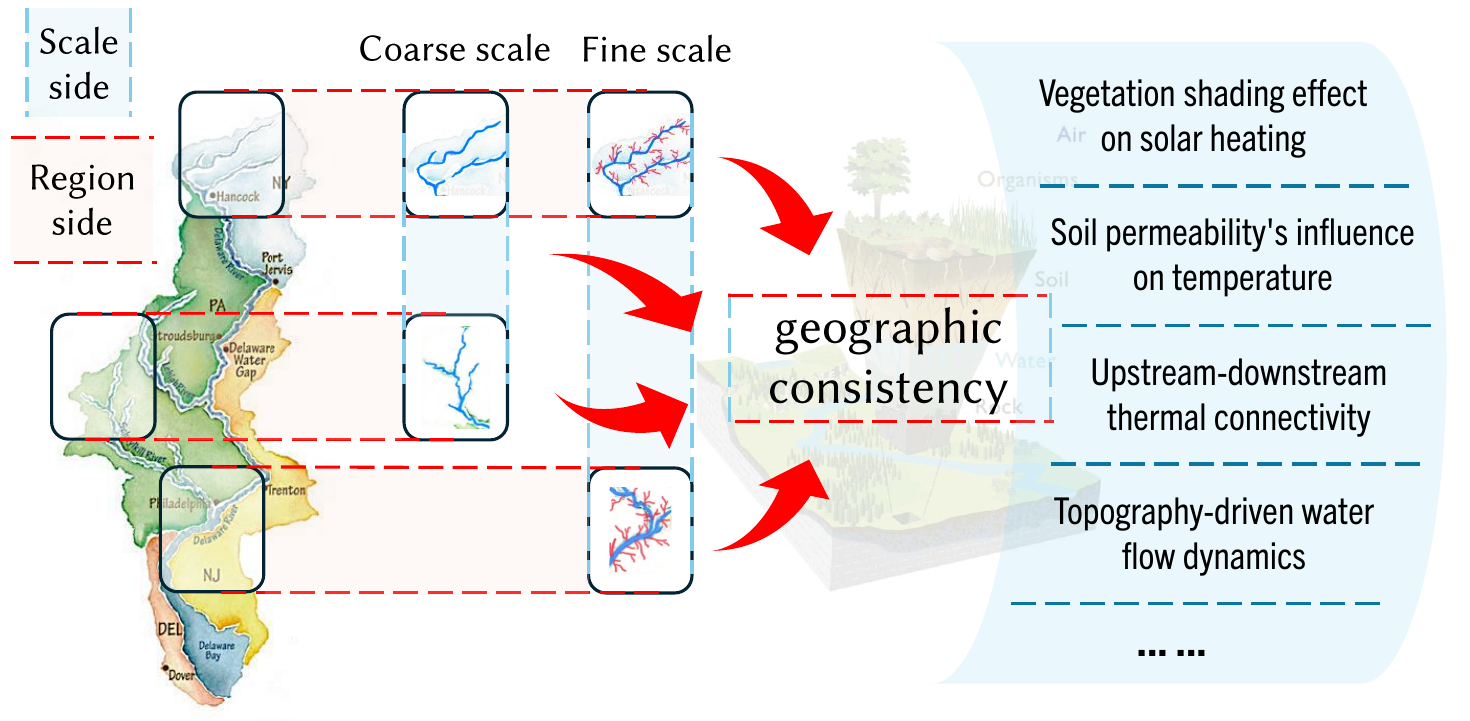} 
\caption{Overview of geographic consistency across different watersheds and at various spatial scales. Different spatial regions on the regional side and both coarse and fine scales on the scale side have shared principles and patterns.}
\label{transfer}
\end{figure}

Despite initial success in isolated prediction tasks, existing methods often fail to make accurate predictions across heterogeneous spatial regions and diverse spatial scales (i.e., different spatial resolutions), which limits their utility in various data analysis tasks and important decision-making processes in managing real environmental ecosystems~\cite{xie2021statistically}. 
In particular, 
observations from different spatial regions exhibit distinct data patterns, due to variations in geological and meteorological conditions, as well as human management. 
As a result, models created from specific regions may not generalize well to other regions. 
On the other hand, model predictions must be provided at the desired spatial scales for specific management tasks. For example, stream water temperature needs to be predicted over discrete river segments of 1 km or shorter to enable effective monitoring and management of fish habitats. Similarly, cloud-resolving models (CRMs) often need to run at a sub-kilometer horizontal resolution to capture important climate patterns~\cite{rasp2018deep}. However, models are typically developed at a coarser resolution where observation data are more available. Directly applying models to finer scales can result in sub-optimal performance when the target system exhibits strong spatial variations.

Ideally, one could build individual models tailored to each specific task, such as a particular region and scale. However, this approach is often impractical because many real environmental ecosystems are sparsely observed or entirely unobserved due to the substantial cost needed for data collection (e.g., in-situ measurements and field surveys), 
especially at fine spatial scales. 
Nevertheless, accurate predictions for these tasks are often as much needed as data-rich tasks for decision-making. 
This raises a challenge: can a model learn from data-rich domains and generalize to data-scarce domains across both regions and scales? 
One promising direction for modeling environmental ecosystems is to leverage underlying geographic consistency on the target variable. 
As shown in Fig.~\ref{transfer}, geographic consistency refers to the shared principles and patterns across different regions and scales, such as the way vegetation shading reduces solar heating, soil permeability affects water temperature, 
slopes guide water flow dynamics, and upstream-downstream thermal connectivity influences energy exchange. These consistent geographic factors drive environmental energy exchange processes regardless of regional or scale differences. 
By incorporating geographic information (e.g., soil properties, vegetation, slope) into the model, the model has a better chance at learning patterns consistent across different spatial regions and spatial scales. 
This raises a critical question: how can we effectively incorporate such geographic information into the model to capture the geo-aware influence of environmental drivers on the target variable dynamics?

\begin{figure*}[t]
\centering
\includegraphics[width=0.95\linewidth]{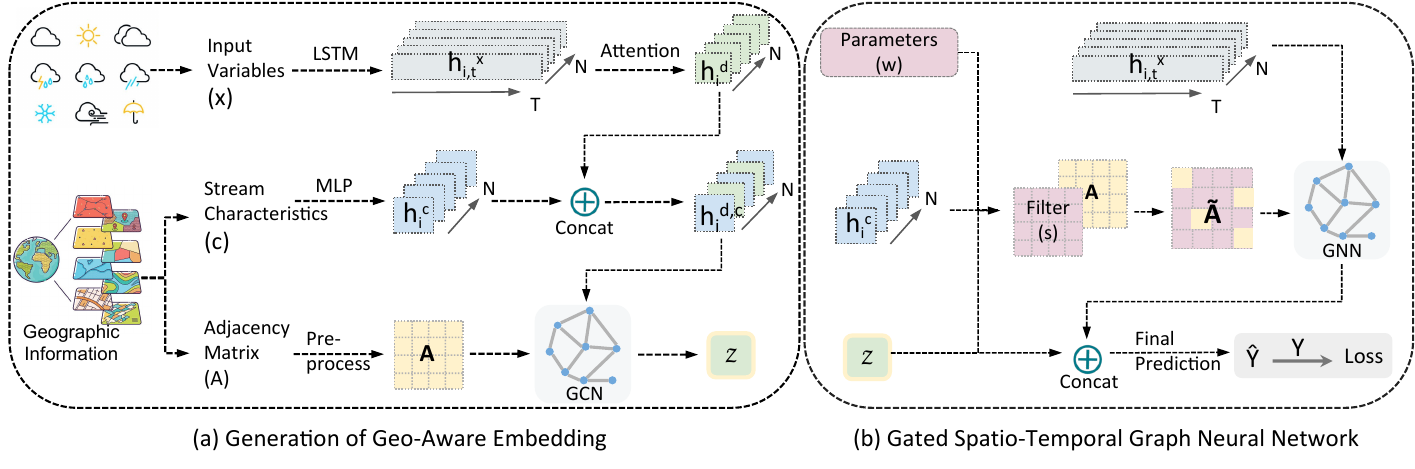} 
\caption{Overview of Geo-STARS framework. The architecture consists of two main components: (a) Generation of Geo-Aware Embedding and (b) Gated Spatio-temporal Graph Neural Network. In (a), input meteorological features $x$, stream characteristics $c$, and stream adjacency matrix $\mathbf{A}$ are processed respectively to produce a geo-aware embedding $z$, which captures the shared underlying geographic principles of stream temperature by embedding overall information from these inputs for each task. In (b), the embedding $z$ and stream characteristics are used to compute adaptive influence filters that modulate the adjacency matrix at each time step. These filters enable intelligent information aggregation from current and past neighboring segments, allowing the model to capture non-uniform spatial dependencies and temporal lags in water temperature dynamics.}
\label{fig:method}
\end{figure*}

In this paper, we aim to develop a geographic-aware model for predicting stream water temperature across different watersheds and at various spatial scales. As a critical factor of water quality, accurate stream temperature prediction is essential for making informed management decisions, maintaining desired aquatic habitats, and understanding related ecological processes. 
We propose a \textbf{Geo}graphic-aware model for \textbf{S}tream \textbf{T}emperature prediction \textbf{A}cross different spatial \textbf{R}egions and \textbf{S}cales (\textbf{Geo-STARS}).
Geo-STARS leverages geographic consistency to enhance the modeling of complex spatio-temporal water dynamics 
across different tasks,
even when these tasks have sparse or no observational data. 
The central idea is to enhance the spatio-temporal modeling of water dynamics by introducing a geo-aware embedding, which is constructed to encode geographic information, including stream characteristics, dynamic meteorological features, and scale information. This embedding enables the model to capture shared geographic principles such as the consistent influence of soil type, soil moisture, and vegetation shading on stream temperature across various tasks. 

Specifically, since stream segments within a watershed are interconnected and can naturally form stream networks, we create a spatio-temporal modeling framework that employs a graph neural network to capture the interactions among streams.  
Time dependencies are also modeled to account for the temporal water dynamics and delayed heat advection across stream segments due to the water travel time. 
More importantly, we enhance this framework using the geo-aware embedding as a filter to selectively modulate the influence of interactions between neighboring river segments. This design is consistent with the fact that the energy exchange between neighboring streams and the resulting temperature changes depend on stream characteristics and local environmental conditions. 
Finally, building on the proposed spatio-temporal framework and geo-aware embedding, we introduce several fine-tuning strategies for efficient model adaptation to new target tasks. The choice of fine-tuning strategies will be based on the availability of observations and stream characteristics for each task.

We evaluate Geo-STARS on real-world datasets of multiple watersheds  
along the eastern coast of the United States
collected over 37 years
at different spatial scales. 
The results show that Geo-STARS outperforms a diverse set of baseline methods in generalizing across different regions and scales, even in data-scarce or entirely unobserved spatial regions. Moreover, GEO-STARTS is shown to perform well even when  geographic characteristics for certain regions are missing. 
The main contributions of this paper can be highlighted as follows:

\begin{itemize}
\item We propose Geo-STARS, a novel geographic-aware model for predicting stream water temperature across watersheds and spatial scales. 
It addresses critical challenges in environmental modeling, including data heterogeneity and observation sparsity, by leveraging shared geographic patterns.

\item Technically, we develop a geo-aware embedding that captures consistent geographic principles across tasks and integrate it into a spatio-temporal graph-based architecture. This embedding guides spatial interactions among river segments and temporal modeling, allowing the framework to intelligently filter interactions between neighboring segments.

\item We introduce effective fine-tuning strategies tailored to varying levels of data availability, including a zero-shot setting. The effectiveness of Geo-STARS in predicting stream water temperature has been demonstrated on multiple watersheds in the United States at varying spatial scales. 
\end{itemize}

\section{Related Work}

\textbf{Modeling of Environmental Ecosystems.} 
Physics-based models have been widely used to simulate complex environmental processes, including different components in water cycles~\cite{hipsey_general_2019,markstrom2012p2s} and plant growth in agroecosystems~\cite{zhou2021quantifying}. However, these models are necessarily approximations of reality and often rely on approximations and parameterizations~\cite{gupta2014debates}.
Consequently, they may exhibit biases even after being calibrated with sufficient training data.
Recent advances in ML offer a great opportunity to improve the modeling of environmental ecosystems by leveraging growing availability of 
data on water, plants, soils, and climate~\cite{yu2025environmental,yu2025foundation}. Previous research has shown the potential of ML-based approaches in modeling agroecosystems~\cite{liu2022kgml_backup,cheng2025knowledge} and freshwater ecosystems~\cite{willard2021predicting,luo2023free,yu2025physics,yu2024adaptive}. 
Importantly, the outcomes of these models can be used to inform critical actions (e.g., water quality treatment, distribution of subsidies~\cite{national2018improving}) to mitigate natural disturbance-incurred food and freshwater shortages, which are essential for continued sustainability and stability. 
Improvements in predicting key variables achieved through the proposed research would significantly benefit these societally relevant decision-making activities. This highlights the critical intersection of environmental science, technology, and policy in addressing global environmental challenges.

\textbf{Spatial Heterogeneity.}
While recent advances in ML and deep learning offer promising ways to mine spatial data, spatial heterogeneity~\cite{atluri2018spatio} – an intrinsic characteristic embedded in spatial data – poses a major challenge as data distributions or generative processes often vary across space at different scales~\cite{xie2021statistically}. “Black-box” ML models, that solely rely on the supervision contained in data, show limited generalizability in scientific problems, especially when applied to out-of-distribution data~\cite{karpatne2024knowledge}. 
Additionally, finer-scale datasets usually need more sample points than typical datasets. However, in many real-world cases, observations are generally available only for limited locations and some specific periods in the environmental field, 
because collecting observations can be expensive, time-consuming, or even unrealistic~\cite{zhuang2020comprehensive}. 
Consequently, most existing methods cannot be applied to other regions or scales beyond the extent of the training data, as the success of standard machine learning models is contingent on having sufficient labeled training data that share a similar distribution with the test data.

\textbf{Transfer Learning.} 
Transfer learning enables leveraging knowledge from data-rich tasks to facilitate the learning for data-scarce tasks (e.g., a new region or a different scale)
~\cite{ma2021transferring}, 
and have shown encouraging results in many environmental problems, including crop yield prediction~\cite{ma2023multisource}, hydrologic prediction~\cite{yao2023can}, flood detection~\cite{zhou2022flooddan}, N2O emission estimation~\cite{liu2022kgml}, and leaf area index estimation~\cite{zhou2023deep}. 
Most existing applications focus on few-shot learning settings, where sparse observations are available in the target domain to fine-tune a pre-trained model~\cite{zhuang2020comprehensive, sadeghi2021structure}. In contrast, zero-shot learning, where no labeled data is available in the target system, presents a significantly harder challenge. In such cases,  models 
can either explicitly reduce co-variate shifts (e.g., using domain adaptation~\cite{zhou2022flooddan}) or leverage 
task-specific contextual inputs, such as environmental conditions and static system characteristics~\cite{kratzert2019towards,xu2023mini}, to enhance generalization. 
However, current works only utilize contextual variables in a simplistic manner, which limits their ability to effectively capture geographically consistent patterns in data-sparse scenarios. Also, existing methods have yet to unify both spatial and scale generalization within a single framework~\cite{yu2025survey,yu2025foundation}.

\textbf{Spatial-Temporal Prediction.}
Spatial-temporal graph modeling is commonly adopted for analyzing the spatial relations and temporal trends of processes within a system~\cite{wu2019graph}. Integrating spatial and temporal dependencies enables a deeper understanding of complex ecosystems and offers a promising way to improve the generalizability of machine learning models~\cite{jia2021physics_sdm,sun2021explore,topp2023stream}. Additionally, Some ML models~\cite{kratzert2019towards,xu2023mini} use meteorological time series data and geographic information (e.g., static stream characteristics) to capture the spatial and temporal dependencies across multiple systems. 
However, the potential connection between spatio-temporal generalizability and the effective use of geographic information remains relatively underexplored. 

\section{Problem Definition}

The goal of this work is to predict stream temperatures by leveraging geographic consistency and spatio-temporal patterns shared across different modeling tasks.  
Each task involves predicting water temperature for a network of connected river segments within a specific watershed and at a specific scale (i.e., low resolution or high resolution based on different hydrological fabric standards). 
Specifically, in our tests, different watersheds are selected from various geographic regions, which span multiple states along the eastern coast of the United States. For each watershed, we create two tasks for modeling at both low and high resolutions, which are defined by the geospatial fabric used for the National Hydrologic Model~\cite{Regan2018NationalHydrologicModel} and the National Hydrography Dataset~\cite{USGS2019NHD}, respectively. The low-resolution (coarse-scale) network contains longer stream segments (with an average length of 10.5 km), while the high-resolution (fine-scale) network consists of much shorter segments (with an average length of 1.3 km) and typically has much fewer observed data points per segment. 

The river networks are represented as a directed graph $\mathcal{G}(\mathcal{V}, \mathcal{E})$, where $\mathcal{V}$ denotes the set of river segments and $\mathcal{E}$ denotes the set of connections among these segments. Specifically, we create an edge $(i, j) \in \mathcal{E}$ if segment $i$ is upstream of segment $j$. The adjacency matrix $\mathbf{A} \in \mathbb{R}^{N \times N}$ represents the adjacency levels between segments, where $N$ is the number of segments. An entry $\mathbf{A}_{ij} = 0$ indicates no edge from segment $i$ to segment $j$, while higher values indicate a shorter river distance between them. More details on generating the adjacency matrix are discussed in the Section~\ref{adj}.


For each river segment $i$, we have access to its input features at multiple daily time steps $x_i = \{{x}_{i,1}, {x}_{i,2}, ..., {x}_{i,T}\}$. These input features include a set of meteorological variables (e.g., solar radiation, rainfall) that are commonly used as drivers for physics-based models. Additionally, we have static stream characteristics for each segment $i$, denoted as 
$c_i$,
which describe the segment's soil, topographic, and geologic characteristics.
Notably, these characteristics contain extensive geographic information that affects the water dynamics, although they may be absent in certain locations due to the cost of measurement. The stream temperature label is represented as 
$\mathbf{Y}=\{{y}_{i,t}\}$, 
where ${y}_{i,t}$ is the water temperature at location $i$ and time step $t$. The observations are only sparsely available for certain time steps and locations.

\section{Method}
In this section, we introduce \textbf{Geo-STARS}, a spatio-temporal graph-based framework designed to predict stream water temperature across different watersheds and spatial scales. The framework captures both geographic knowledge and spatio-temporal 
dynamics for effective generalization across diverse tasks. 
As shown in Fig.~\ref{fig:method}, Geo-STARS consists of two key components: (1) a \textit{geo-aware embedding generation} module that captures shared geographic principles across regions and scales, and (2) a \textit{gated spatio-temporal graph neural network} that models spatial interactions among stream segments and temporal evolution of water temperature.


\subsection{Generation of Geo-Aware Embedding}



Environmental systems have intrinsic spatial heterogeneity across different watersheds and spatial scales. 
To effectively generalize across these tasks, our model must encode consistent geographic principles that govern stream temperature dynamics.  Hence, we construct a geo-aware embedding $z$ to encode the underlying consistent geographic nature for each task. 
The geo-aware embedding has two major purposes: (1) it modulates the influence between spatially connected stream segments, and (2) it captures overall temporal trends by emphasizing time periods that most strongly reflect the distribution of water temperature. 

This embedding encodes information from multiple sources for each task: (1) local meteorological features, which influence temperature dynamics over time; (2) stream segment characteristics, which capture static geographic and hydrologic properties; and (3) scale information, which reflects the spatial resolution and connectivity within the stream network. 
Specifically, 
we hypothesize that local meteorological data can indicate the overall trends of water temperature dynamics for a specific task.  
Thus, we extract temporal patterns from meteorological inputs $\{x_{i,t}\}$ (e.g., air temperature, rainfall) and   
aggregate them across the stream network based on spatial interactions. 
Additionally, we process the static stream characteristics, which capture the task-specific stream and hydrological conditions,  
and integrate them with the meteorological patterns. 
Finally, we include scale information by incorporating stream distances into the spatial aggregation process, ensuring the embedding is scalable and adaptable to various regions and scales.
The complete embedding $z$ is then used throughout the model to guide both spatial interactions and temporal dependencies. In the following, we will elaborate on the procedure for generating the geo-aware embedding.

\subsubsection{\textbf{Embedding Temporal Meteorological Data and Stream Characteristics}}
We use the long short-term memory (LSTM) network~\cite{hochreiter1997long} to capture the temporal patterns of meteorological data, as the weather often exhibits important short-term and long-term variations that influence stream temperature.
Specifically, for each river segment $i$, the time series of meteorological features $\{x_{i,t}\}_{t=1}^T$ is processed through the LSTM to produce temporal feature embeddings $\{h^{\text{x}}_{i,t}\}_{t=1}^T$.

We then apply a temporal attention mechanism~\cite{vaswani2017attention} to transform these embeddings into a fixed-length representation (independent of sequence length $T$).
This allows the model to focus on the most informative time periods for overall stream temperature trends.  
The attention weights $\alpha_{i,t}$ for each time step are computed as:
\begin{equation}
\small
\begin{aligned}
\tilde{\alpha}_{i,t} &= h^{\text{x}}_{i,t} \cdot q_h, \\
\alpha_{i,t} &= \frac{\exp(\tilde{\alpha}_{i,t})}{\sum_{t'=1}^T \exp(\tilde{\alpha}_{i,t'})},
\end{aligned}
\end{equation}
where $q_h$ is a learnable query vector.

The final fixed-length meteorological embedding $h^{\text{d}}_i$ for segment $i$ is computed as a weighted sum of the LSTM outputs:
\begin{equation}
\small
h^{\text{d}}_i = \sum_{t=1}^T \alpha_{i,t} h^{\text{x}}_{i,t}.
\end{equation}

Next, we incorporate static geographic characteristics. Each segment's stream characteristics $c_i$ are passed through a multi-layer perceptron (MLP) to obtain a latent representation $h_i^c$:
\begin{equation}
\small
\begin{aligned}
h_i^c &= \text{MLP}(c_i), \\
h_i^{d,c} &= [h_i^d, h_i^c],
\end{aligned}
\label{eq:mlp}
\end{equation}
where $h_i^{d,c}$ is the concatenated vector of dynamic (meteorological) and static (stream) embeddings for segment $i$.

\subsubsection{\textbf{Spatial Aggregation}}
\label{adj}

To capture the scale information and spatial dependencies among river segments, we utilize the graph adjacency matrix 
to aggregate dynamic and static feature representations. Specifically, 
we first generate the adjacency matrix
based on the stream distances between river segment outlets, denoted as $\text{dist}(i,j)$ for each pair of segments $i$ and $j$.  
We calculate a unified mean and standard deviation from multiple tasks of varying scales and standardize the stream distance using these values: 
\begin{equation}
\small
   \mathbf{D}_{ij} = \frac{\text{dist}(i, j)-\text{mean}_{i,j}\{\text{dist}(i,j)\}}{\text{std}_{i,j}\{\text{dist}(i,j)\}}. 
\end{equation}

The  $\{\mathbf{D}_{ij}\}$  values obtained can reflect the relative scale difference between different tasks.
Then we reverse distance measures to create the adjacency matrix  as:
\begin{equation}
\small
   \mathbf{A}_{ij} = 1/(1 + \text{exp}(\mathbf{D}(i, j)))
\end{equation}

This transformation is chosen because it 
assigns similarly low weights for segments that are extremely far away, which have little influence on the water temperature dynamics.

Using the obtained adjacency matrix, we aim to aggregate meteorological and stream embeddings of all nodes into a task-specific graph-level embedding. First, we use a graph convolutional layer to aggregate features from neighboring segments. The hidden representation $g_i$ for each segment $i$ is computed as:
\begin{equation}
\small
g_i = \text{ReLU} \left( \sum_{j \in \mathcal{N}(i)} \mathbf{A}_{ij} \, h^{\text{d,c}}_j \right),
\label{eq:GCN}
\end{equation}
where $\mathcal{N}(i)$ denotes the set of neighboring segments of segment $i$, including $i$ itself. 

Next, to generate a compact, task-level embedding that is independent of the number of segments,  we perform an attention-based graph pooling to embed the entire network. This allows the embedding to be scalable and adaptable to various regions and scales. The attention weight $\beta_i$ for node $i$ is computed as:
\begin{equation}
\small
\begin{aligned}
\tilde{\beta}_i &= g_i \cdot q_g, \\
\beta_i &= \frac{\exp(\tilde{\beta}_i)}{\sum_{j=1}^{N} \exp(\tilde{\beta}_j)},
\end{aligned}
\end{equation}
where $q_g$ is a learnable global query vector.

The final geo-aware embedding $z$ for the task is then obtained by a weighted sum of the node embeddings:
\begin{equation}
\small
z = \sum_{i=1}^{N} \beta_i g_i.
\end{equation}

\subsection{Gated Spatio-Temporal Graph Neural Network}

Given the interconnected nature of stream networks, effective modeling requires capturing the influence of neighboring river segments. However, the strength of these influences is not uniform, so modeling must not only depend on static spatial adjacency but also consider the dynamic nature of water flow.
To this end, we adopt a graph neural network (GNN)-based architecture in which each stream segment is represented as a node, and edges represent hydrological connections based on stream distances. The major issue with standard GNNs is that the aggregation operation weighs neighbors solely based on the adjacency values (i.e., stream distances). However, the influence can vary significantly for segments with similar stream distances due to differences in stream characteristics (e.g., width, depth, and slope,  which determine streamflow). 

To address this issue, we propose a spatial gating mechanism that dynamically modulates the influence of neighboring segments. For each stream segment, this mechanism incorporates: (1)~local stream characteristics, (2)~the task-level information from the geo-aware embedding $z$, and (3)~scale-aware stream distance information to its neighbors. By learning to gate neighbor contributions based on these information sources, the model can better differentiate which neighboring segments have stronger or weaker influence and capture relationships that align with underlying water dynamics. 

In addition, 
stream temperature is affected by 
delayed responses to upstream conditions due to water travel time. To capture this, we 
extend the GNN to capture lagged effects of segment interactions in water dynamics.  Therefore, the model considers both current and past representations of neighboring nodes.

Together, 
the gating mechanism allows our framework to 
model the complex, non-uniform, and time-varying dependencies in stream temperature prediction.

\subsubsection{\textbf{Spatial Gating Mechanism}}
To enhance the model’s ability to capture non-uniform spatial dependencies, we introduce a spatial gating mechanism that computes a pairwise influence filter $s_{ij}$ between each node $i$ and its neighbor $j$. This filter is based on the similarity between their stream characteristics, modulated by the geo-aware embedding $z$, specifically as follows:
\begin{equation}
\small
s_{ij} = w_{ij} \langle\, h_i^c\odot z, h_j^c\odot z \rangle,   
\label{filter1}
\end{equation}
where $\odot$ denotes element-wise multiplication, $\langle\, , \rangle$ is the inner product, and $w \in \mathbb{R}^{N \times N}$ is a trainable weight matrix. This filter enables the model to weigh the spatial influence of each neighbor based on multiple elements rather than just connectivity.

We then use these influence filters to adaptively reweight the adjacency matrix. Each entry $\mathbf{A}_{ij}$ is modulated by its corresponding filter $s_{ij}$ and then perform row-wise normalization via a softmax function:
\begin{equation}
\small
\tilde{\mathbf{A}}_{i.} = \text{softmax}(\mathbf{A}_{i.}\odot [s_{i1},s_{i2},..,s_{iN}]),
\end{equation}
where $\mathbf{A}_{i\cdot}$ denotes the $i^{\text{th}}$ row of the adjacency matrix $\mathbf{A}$.

\subsubsection{\textbf{Delayed Temporal Gating Mechanism}}

Meanwhile, we apply a similar gating mechanism to capture temporal dependencies. From the temporal perspective, the water dynamics for each segment depend on (1) its own state at the previous time step due to the heat transfer process, and (2) the status (e.g., weather conditions and water temperature) of neighboring streams at the previous time step, due to the advected heat and the travel time of water flows across segments. Hence, we construct another set of influence filters $\{s'_{ij}\}$ using separate parameters $w'$, allowing the model to learn delayed dependencies:
\begin{equation}
\small
\begin{aligned}
s'_{ij} &= w'_{ij} \langle\, h_i^c\odot z, h_j^c\odot z \rangle \\
\tilde{\mathbf{A}}'_{i.}&= \text{softmax}(\mathbf{A}_{i.}\odot [s'_{i1},s'_{i2},..,s'_{iN}])
\label{filter2}
\end{aligned}
\end{equation}

\begin{figure*}[t]
\centering
\small
\includegraphics[width=0.8\linewidth]{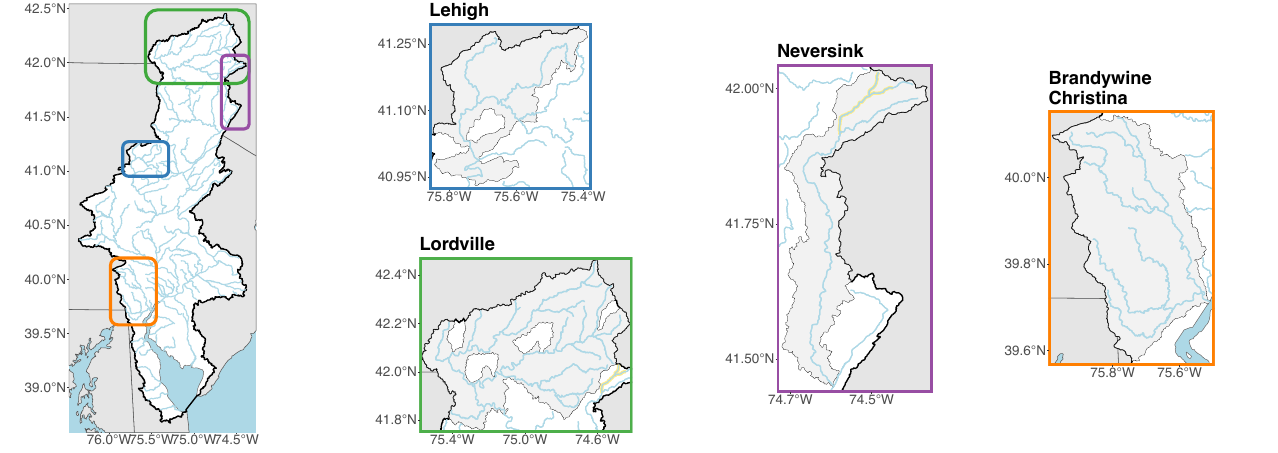} 
\caption{Map of four distinct and ecologically varied watersheds in the eastern United States.}
\label{fig:basins}
\end{figure*}

\subsubsection{\textbf{Final Prediction}}
For each segment $i$, we first utilize the LSTM to extract temporal patterns from input features as $\{h_{i,t}\}_{t=1}^T = \text{LSTM}(\{x_{i,t}\}_{t=1}^T)$. 
Then using the gated adjacency matrices $\tilde{\mathbf{A}}$ and $\tilde{\mathbf{A}}'$, we aggregate information from neighboring nodes at both the current time step $t$ and the previous time step $t{-}1$. 
The aggregated representation is computed as:
\begin{equation}
\small
\begin{aligned}
\hat{h}^{\text{x}}_{i,t} &= \sum_{j \in \mathcal{N}(i)}^{N} \tilde{\mathbf{A}}_{i,j} \cdot h^{\text{x}}_{j,t} \\
\hat{h}^{\text{x}}_{i,t-1} &= \sum_{j \in \mathcal{N}(i)}^{N} \tilde{\mathbf{A}}'_{i,j} \cdot h^{\text{x}}_{j,t-1} 
\end{aligned}
\end{equation}

Finally, we concatenate the aggregated representations from both time steps along with the geo-aware embedding $z$, and pass them through a multi-layer perceptron (MLP) to generate the temperature prediction:
\begin{equation}
\small
\begin{gathered}
\hat{y}_{i,t} = \text{MLP} \left( [\hat{h}^{\text{x}}_{i,t}, \hat{h}^{\text{x}}_{i,t-1}, z] \right)
\end{gathered}
\end{equation}
This design allows the model to integrate fine-grained spatial influences, temporal lags, and high-level geographic context in a unified prediction pipeline.


In summary, the gated spatio-temporal GNN leverages stream-level characteristics, scale-level interactions, and shared geographic representations to model complex spatio-temporal dependencies. This design enhances the model's ability to capture patterns that are consistent with underlying water dynamics and generalizable over space and time. 

\subsection{Model Generalization to New Tasks}

After being pre-trained on a collection of source tasks, Geo-STARS can be directly applied to a new task (e.g., a different watershed or a new spatial scale within a known watershed) in a zero-shot manner. This is particularly valuable in scenarios where no observation data is available for the new task, enabling prediction in completely unmonitored regions. This generalizability comes from the design of the geo-aware embedding $z$, which is computed entirely from input features such as meteorological data, stream characteristics, and spatial structure—none of them require observations. As a result, Geo-STARS can leverage previously learned geographic principles to guide its predictions without adaptation.

The model can be further fine-tuned when observation data for the new task becomes available. In practice, the collected characteristics may be noisy or incomplete (i.e., missing relevant characteristics), and thus may not precisely capture the geo-aware information needed for modulating water dynamics. Moderate fine-tuning can address this gap and adapt more effectively to the target task while keeping the knowledge learned from the source tasks.

We provide three fine-tuning strategies, each tailored to different levels of data availability and task conditions:

\begin{itemize}
    \item \textbf{Complete Fine-tuning (Complete)}: This strategy updates all model parameters using the available observation data from the new task. It is best suited when both observation data and stream characteristics are sufficiently available, allowing the model to fully adapt to the new environment.

    \item \textbf{Partial Fine-tuning (Geo-Related)}: This strategy updates only the model components associated with stream characteristics (i.e, the MLP defined in~\cref{eq:mlp}), while keeping all other parameters fixed. This reduces the risk of overfitting and is more appropriate when observation data is limited but characteristic information is reliable.

    \item \textbf{Geo-Aware Embedding Fine-tuning (Geo-Focus)}: This strategy keeps all model parameters frozen and fine-tunes only the geo-aware embedding $z$, treating it as a task-specific prompt to better reflect the new task's overall geographic context. It is particularly useful when observation data is extremely scarce or when stream characteristics are missing or unreliable for the new task.
\end{itemize}

\begin{table}[t!]
\small
\centering
\caption{Dataset statistics}
\rowcolors{2}{gray!10}{white}
\begin{tabular}{c|c|c|c}
\toprule
\textbf{\footnotesize Datasets} & \makecell{\footnotesize \textbf{\#Segments}} & \makecell{\footnotesize \textbf{\#Segments} \\ \footnotesize  \textbf{Without} \\ \footnotesize \textbf{Observation}} & \footnotesize \makecell{{\textbf{Missing}} \\ \footnotesize \textbf{Observations(\%)}} \\
\midrule
\textbf{\footnotesize NS$_\text{c}$}    & 13 & 0 & 78.02 \\
\midrule
\textbf{\footnotesize NS$_\text{f}$}    & 63 & 27 & 93.42 \\
\midrule
\textbf{\footnotesize UL$_\text{c}$}    & 8 & 0 & 94.76 \\
\midrule
\textbf{\footnotesize UL$_\text{f}$}    & 83 & 67 & 99.49 \\ 
\midrule
\textbf{\footnotesize LV$_\text{c}$}    & 56 & 11 & 87.80 \\  
\midrule
\textbf{\footnotesize LV$_\text{f}$}    & 319 & 222 & 97.38 \\  
\midrule
\textbf{\footnotesize CRW$_\text{c}$}   & 42 & 10 & 88.34 \\ 
\midrule
\textbf{\footnotesize CRW$_\text{f}$}   & 174 & 37 & 97.11  \\  
\bottomrule
\end{tabular}
\label{table:dataset}
\end{table}

\begin{table}[t!]
\small
\centering
\caption{Experiment names with their source and target tasks.}
\begin{tabular}{c|c|c}
\toprule
\textbf{Names} & \textbf{Source tasks} & \textbf{Target tasks} \\
\midrule
R1    & \makecell[c]{NS$_\text{c}$, NS$_\text{f}$, CRW$_\text{c}$, CRW$_\text{f}$, LV$_\text{c}$, LV$_\text{f}$} & UL$_\text{c}$, UL$_\text{f}$ \\
\midrule
R2    & \makecell[c]{UL$_\text{c}$, UL$_\text{f}$, CRW$_\text{c}$, CRW$_\text{f}$, LV$_\text{c}$, LV$_\text{f}$} & NS$_\text{c}$, NS$_\text{f}$ \\
\midrule
R3    & \makecell[c]{NS$_\text{c}$, NS$_\text{f}$, UL$_\text{c}$, UL$_\text{f}$, CRW$_\text{c}$, CRW$_\text{f}$} & LV$_\text{c}$, LV$_\text{f}$ \\
\midrule
R4    & \makecell[c]{NS$_\text{c}$, NS$_\text{f}$, UL$_\text{c}$, UL$_\text{f}$, LV$_\text{c}$, LV$_\text{f}$} & CRW$_\text{c}$, CRW$_\text{f}$ \\ 
\midrule
S1    & \makecell[c]{NS$_\text{c}$, UL$_\text{c}$, CRW$_\text{c}$, CRW$_\text{f}$, LV$_\text{c}$, LV$_\text{f}$}& NS$_\text{f}$, UL$_\text{f}$ \\  
\midrule
S2    & \makecell[c]{NS$_\text{c}$, NS$_\text{f}$, UL$_\text{c}$, UL$_\text{f}$, CRW$_\text{c}$, LV$_\text{c}$} & CRW$_\text{f}$, LV$_\text{f}$ \\ 
\midrule
S3    & \makecell[c]{NS$_\text{c}$, UL$_\text{c}$, CRW$_\text{c}$, LV$_\text{c}$, LV$_\text{f}$} & \makecell[c]{NS$_\text{f}$, CRW$_\text{f}$, UL$_\text{f}$} \\ 
\midrule
S4    & \makecell[c]{NS$_\text{c}$, NS$_\text{f}$, UL$_\text{c}$, CRW$_\text{c}$, LV$_\text{c}$} & \makecell[c]{CRW$_\text{f}$, UL$_\text{f}$, LV$_\text{f}$} \\  
\bottomrule
\end{tabular}
\label{group_names}
\end{table}

\section{Experiments}

In this section, we provide a comprehensive assessment of the proposed methods for different new tasks. Our primary emphasis is on 
the performance evaluation under data-sparse scenarios and even in the absence of available labels, which is a common issue for modeling real environmental ecosystems. 
We also design an ablation study to demonstrate the effectiveness of each component in the proposed method. 
Finally, we conduct additional experiments to test the generalizability of the proposed method when certain stream characteristics are unavailable in target tasks.

\begin{table*}[t]
\centering
\caption{Comparison of average RMSE for stream water temperature prediction under the sparsity level of 0.1\%. The best results are \textbf{bold}, while the second-best results are \underline{underlined}. The second column shows different fine-tuning strategies.}
\begin{tabular}{l|p{1.8cm}|c|c|c|c|c|c|c|c}
\toprule
\textbf{Methods} & \textbf{Strategies} & \textbf{R1} & \textbf{R2} & \textbf{R3} & \textbf{R4} & \textbf{S1} & \textbf{S2} & \textbf{S3} & \textbf{S4} \\
\midrule
\textbf{LSTM} & Complete & 3.04 & 2.87 & 3.19 & 3.09 & 2.85 & 3.25 & 3.01 & 2.97 \\
\textbf{Transformer} & Complete & 3.43 & 2.76 & 3.42 & 2.78 & 3.06 & 3.04 & 2.74 & 3.16 \\
\textbf{RGRN} & Complete & 2.12 & 2.48 & 3.49 & 3.17 & 2.33 & 3.26 & 2.88 & 3.49 \\
\textbf{GCN-LSTM} & Complete & 3.03 & 3.12 & 3.95 & 3.77 & 3.41 & 3.54 & 3.48 & 4.02 \\
\textbf{Mini-Batch} & Complete & 2.11 & 2.75 & 3.39 & 3.25 & 2.53 & 3.84 & 2.51 & 3.48 \\
\textbf{Graph-WaveNet} & Complete & 2.74 & 3.35 & 3.87 & 3.60 & 3.05 & 3.35 & 2.98 & 3.34 \\
\textbf{EA-LSTM} & Complete & 1.99 & 2.27 & 3.23 & 2.72 & 2.37 & 3.16 & 2.61 & 2.74 \\
\textbf{SR-MTL} & Complete & 2.15 & 2.63 & 3.20 & \underline{2.09} & 2.75 & 2.91 & 2.36 & 2.84 \\
\textbf{STCGAN} & Complete & 2.12 & 2.07 & 3.15 & 2.59 & 3.49 & 3.04 & 2.57 & 3.38 \\
\cmidrule{1-10}
\multirow{3}{*}{\textbf{Geo-STARS}} & Complete & \textbf{1.57} & \textbf{1.98} & \textbf{2.80} & \textbf{2.08} & \underline{1.63} & \textbf{2.59} & \textbf{1.73} & \textbf{2.17} \\
& Geo-Related & \underline{1.80} & 2.12 & \textbf{2.80} & 2.23 & 1.64 & \underline{2.79} & 1.95 & \underline{2.32} \\
& Geo-Focus & 1.87 & \underline{2.03} & 3.88 & 2.76 & \textbf{1.60} & 3.13 & \underline{1.94} & 2.42 \\
\bottomrule
\end{tabular}
\label{finetune_sparsity}
\end{table*}

\subsection{Experimental Setup}

\subsubsection{\textbf{Dataset Description}}

The eight different water temperature datasets are collected from four geographically distinct and ecologically varied watersheds located along the east coast of the United States, as shown in Fig.~\ref{fig:basins}.
In particular, we represent the four watersheds as Neversink (\textbf{NS}), Upper Lehigh (\textbf{UL}), {Lordville} (\textbf{LV}), and Brandywine-Christina (\textbf{CRW}). Each watershed has both input drivers and target water temperature observations at two different spatial scales. Specifically,
for each of the four datasets, we create a low-resolution dataset and a high-resolution dataset. The coarse-scale river segments (coarse level, or \textbf{c}) were defined by the geospatial fabric used for the National Hydrologic Model~\cite{regan2018description} and have an average segment length of 10.5 km in the Delaware River Basin. In contrast, the fine-scale segments (fine level, or \textbf{f}) were defined by the National Hydrography Dataset~\cite{USGS2019NHD} and have an average segment length of 1.3 km.
The coarse-scale networks \textbf{NS$_\text{c}$}, \textbf{UL$_\text{c}$}, \textbf{LV$_\text{c}$}, and \textbf{CRW$_\text{c}$} contain 13, 8, 56, and 42 river segments, respectively, while their corresponding fine-scale networks \textbf{NS$_\text{f}$}, \textbf{UL$_\text{f}$}, \textbf{LV$_\text{f}$}, and \textbf{CRW$_\text{f}$} contain 63, 83, 319, and 174 segments, respectively.

The water temperature observation data are pulled from the U.S. Geological Survey's National Water Information System~\cite{usgeologicalsurvey} and the Water Quality Portal~\cite{waterquality}, the largest standardized water quality dataset for inland and coastal waterbodies. 
While daily average water temperature observations are available for certain segments, they are not consistently available for every date. 
The number of temperature readings for each segment varies and could differ significantly across datasets. Some segments have no recorded water temperature observations, but we retain them to preserve the completeness of the stream networks and to use their input features in modeling their influence to other segments. 
For some datasets, the segments with no observations constitute a large portion, e.g., UL$_\text{f}$, LV$_\text{f}$. The proportion of missing observations is significant across all datasets, especially at fine spatial scales, the ratio can be over 90\%. 
Table~\ref{table:dataset} provides a summary of dataset statistics.

On a daily scale, we incorporate 7 basic meteorological features (daily average air temperature, solar radiation, rainfall, and potential evapotranspiration). We also provide 65 static stream characteristics for each segment in the river networks (e.g., soil characteristics, vegetation, snow area, elevation, and slope). These variables may not always be available for every stream segment and its local area 
due to the difficulty in collecting these characteristics.

This dataset spans from October 1, 1979, to September 30, 2021, covering 37 years (15,341 days). We train our model using data from October 01, 1984, to September 30, 2005 and then test the model with data from October 01, 1979, to September 30, 1984, from October 01, 2010, to September 30, 2015, and from October 01, 2020 to September 30, 2021, covering a total of 11 years.

\begin{figure*}[t!]
\centering
\includegraphics[width=\linewidth]{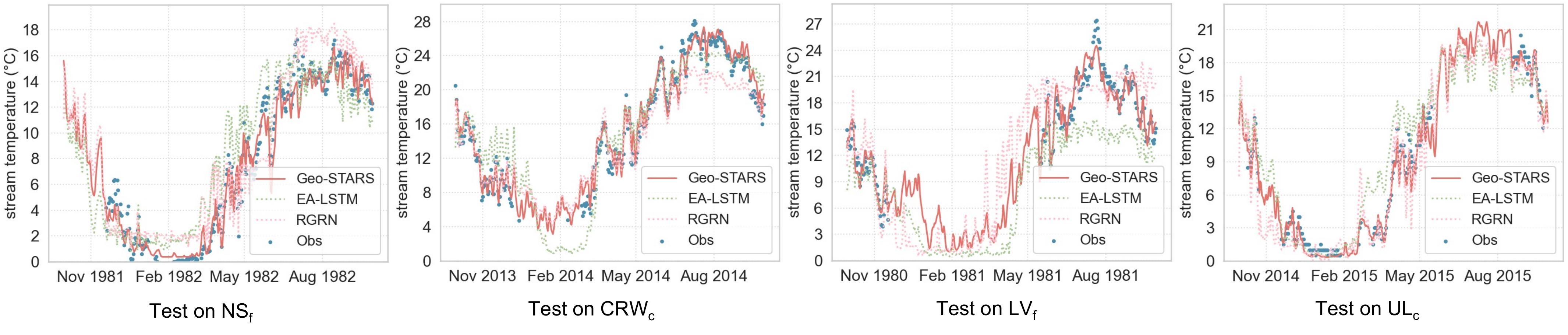} 
\caption{Time-series comparison of stream water temperature predictions by Geo-STARS, EA-LSTM and RGRN.}
\label{pred_obs}
\end{figure*}

\subsubsection{\textbf{Implementation Details}}
We implemented our methods using PyTorch == 2.0.1, on an RTX 4090 GPU, and employed an ADAM~\cite{kingma2014adam} optimizer with a 0.1 learning rate for training, and 0.01 for fine-tuning on 0.1\% sparsity level. Additionally, 
the hidden dimensions of input features and characteristics are 7 and 128, respectively.


\subsubsection{\textbf{Evaluation Metrics}}
Following~\cite{jia2021physics_sdm}, we use Root Mean Square Error (RMSE) to evaluate all methods.
Specifically, RMSE quantifies the square root of the average squared differences between the predicted and observed stream temperatures. It is a widely used indicator for regression accuracy in environmental modeling. A lower RMSE indicates better predictive accuracy.

\subsubsection{\textbf{Baselines}}

We compare the performance of Geo-STARS with seven state-of-the-art baselines developed for stream modeling, including EA-LSTM~\cite{kratzert2019towards}, GCN-LSTM~\cite{sun2021explore}, Mini-Batch~\cite{xu2023mini}, Transformer~\cite{vaswani2017attention}, RGRN~\cite{jia2021physics_sdm}, Graph-WaveNet~\cite{topp2023stream}, SR-MTL~\cite{chen2023meta}, STCGAN~\cite{kalanat2024spatial} and the basic LSTM model~\cite{shen2021applications}, which is widely used in environmental modeling tasks. 
Among these methods, SR-MTL is a meta transfer learning method, and STCGAN~\cite{kalanat2024spatial} is an adversarial network for multi-source domain adaptation. GCN-LSTM, Graph-WaveNet, and RGRN are spatio-temporal models that incorporate meteorological features and adjacency matrices. In contrast, Mini-Batch, and EA-LSTM use meteorological features and stream characteristics originally in their studies but lack mechanisms to integrate adjacency information. To ensure a fair comparison, we include segment characteristics as additional inputs for the baselines that do not natively utilize them. 

While other spatio-temporal methods exist, 
not all methods are capable of incorporating meteorological features, stream characteristics, and adjacency matrix simultaneously. Therefore, our selection of baselines focuses on methods that either natively support these inputs or have been tested in similar water prediction tasks.

\begin{figure}[t!]
\centering
\small
\includegraphics[width=\columnwidth]
{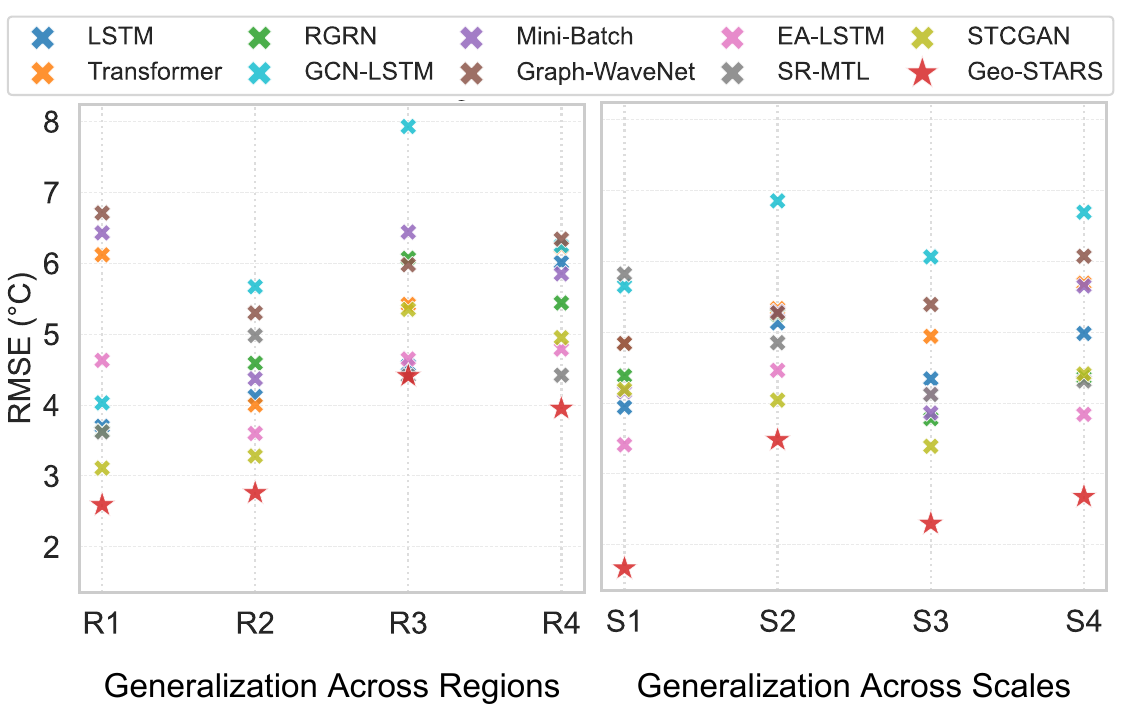}
\caption{Comparison of average RMSE for stream water temperature prediction in zero-shot setting. }
\label{zero_shot}
\end{figure}

\begin{figure*}[t!]
\centering
\includegraphics[width=0.95\linewidth]
{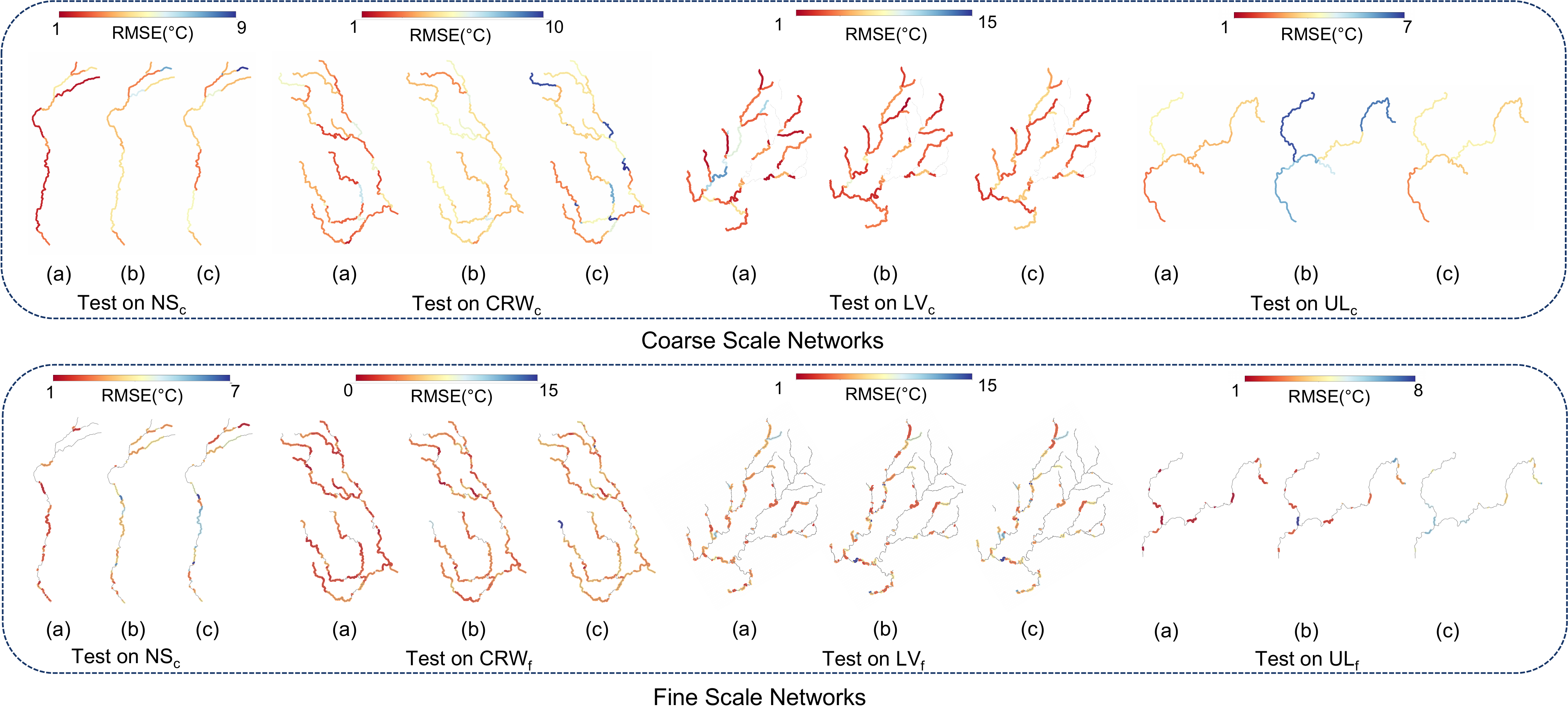} 
\caption{RMSE values for stream water temperature prediction across different segments in all test regions at both coarse and fine spatial scales: \textbf{NS$_\text{c}$}, \textbf{CRW$_\text{c}$},
\textbf{LV$_\text{c}$}, \textbf{UL$_\text{c}$}, \textbf{NS$_\text{f}$}, \textbf{CRW$_\text{f}$},
\textbf{LV$_\text{f}$}, and \textbf{UL$_\text{f}$}. Each group of three maps corresponds to results from different models: (a) Geo-STARS, (b) EA-LSTM, and (c) RGRN. Colors indicate RMSE values for each segment, with red representing lower RMSE (better performance) and blue representing higher RMSE (worse performance). Gray segments in the fine-scale networks denote areas with no available observations, where RMSE could not be computed.} 
\label{rmse_map_CRW}
\end{figure*}

\subsection{Main Results}

We evaluate Geo-STARS on eight experimental settings involving different training and testing tasks. Detailed descriptions of these experimental configurations are provided in Table~\ref{group_names}. The first four experiments (named R1–R4) test model generalizability to new watersheds across distinct geographic regions, with coarse-scale and fine-scale modeling tasks available in both training and testing tasks. The remaining four experiments (named S1–S4) focus on fine-scale predictions for watersheds whose coarse-scale counterparts are seen during training.

\subsubsection{Few-shot Setting: Limited Observations in New Tasks. }
Table~\ref{finetune_sparsity} shows the average predictive performance for all methods under the sparsity level of 0.1\%. It is obvious that Geo-STARS consistently achieves the best or second-best performance across all eight experiments. It shows the superiority of Geo-STARS in generalizing to various tasks with limited observations by effectively leveraging geographic consistency underlying stream characteristics and scale information in the model architecture. 
We can also observe different performances when we adopt different fine-tuning strategies. The complete fine-tuning strategy achieves the best performance in most cases, as it allows the model to adapt fully to target tasks. Geo-Related and Geo-Focus strategies also offer competitive performance in general, while keeping the knowledge learned from the training tasks. 
All these proposed strategies capture important stream characteristics, and the model can be adjusted properly using available observations. 
The detailed experimental results of each target task are provided in the appendix.

To clearly show the superior performance, Fig.~\ref{pred_obs} shows the predicted water temperature over a one-year period for different watersheds and scales. It shows that the predictions made by Geo-STARS align better with true observations compared to the other two strong baselines, especially in winter and summer time. 
The predictions by the EA-LSTM model, while reasonably tracking the general trend, fall short in replicating the lower and higher temperatures and recognizing the subtle changes. The RGRN model creates smooth predictions, overlooking peaks and valleys that Geo-STARS consistently identifies and traces. 

\subsubsection{Zero-shot Setting: No Observation in New Tasks. }
Fig.~\ref{zero_shot} shows the average predictive performance in the zero-shot setting, where no observations are used for adaptation towards target tasks. Geo-STARS outperforms all baselines across all experiments, especially for the generalization across scales. 
The geo-aware embedding effectively transfers knowledge learned from source tasks by capturing shared geographic principles. This allows the model to make accurate predictions even when no observations are available, as long as stream characteristics and the adjacency matrix are provided. 
The detailed RMSE performance for each target task is provided in the appendix.

Fig.~\ref{rmse_map_CRW} shows the zero-shot RMSE performance of stream temperature predictions across different river segments in all watersheds NS, CRW, LV, UL, at both coarse ($\text{c}$) and fine ($\text{f}$) spatial scales. Geo-STARS outperforms the 
baselines in almost all the segments, achieving lower RMSE values without any fine-tuning. In contrast, baseline methods often exhibit high errors in some segments due to their inability to adjust to new regions without fine-tuning. Geo-STARS overcomes this limitation by effectively leveraging the geographic consistency and spatial structure. 

\begin{figure}[t!]
\centering
\small
\includegraphics[width=\columnwidth]
{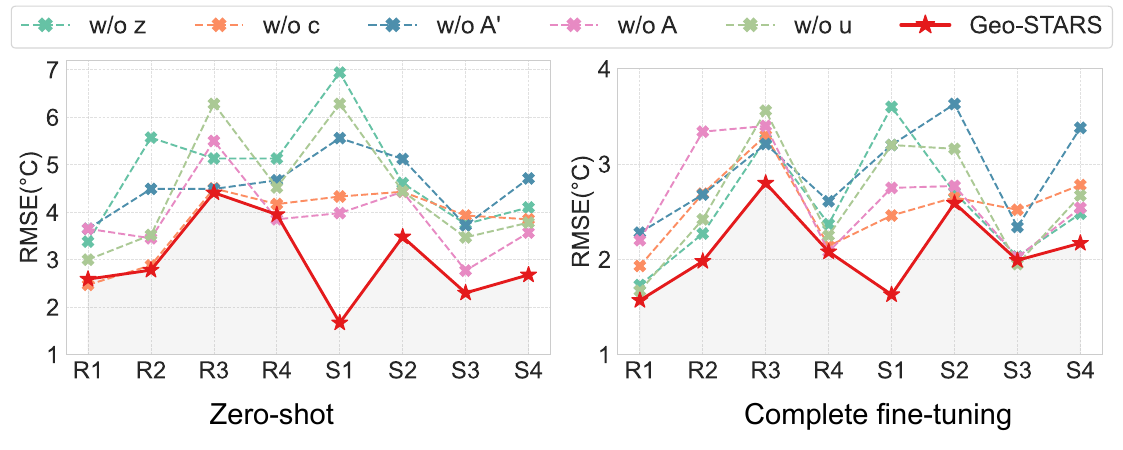}
\caption{Results of ablation study, which demonstrates the impact of different components on the overall performance of our model. The left plot represents results in zero-shot setting, the right plot shows complete fine-tuning results.}
\label{ablation}
\end{figure}

\subsection{Ablation Studies}

In this section, we conduct ablation studies on all experiments to demonstrate the effectiveness of our key modules, including the geo-aware embedding, the gating mechanism, and the method used for processing the adjacency matrix. We use a set of Geo-STARS variants, as follows:
\begin{itemize}
\item \textbf{w/o z}: The geo-aware embedding $z$ is removed from all the processes in the model.
\item \textbf{w/o c}: The influence filters $s_{ij}$ and $s'_{ij}$ in the spatial and temporal gating mechanisms (see Eq.~\eqref{filter1} and~\eqref{filter2}) are simplified to compute without using stream characteristics $h_i^c$ and $h_j^c$.
\item \textbf{w/o A}': The $\tilde{\mathbf{A}}'_{i}$ is removed, ignoring the 
gating mechanism for the delayed influence from the previous time step $t-1$.
\item \textbf{w/o A}: The $\tilde{\mathbf{A}}_{i}$ is removed, ignoring the 
gating mechanism for the spatial interactions at the current time step $t$.
\item \textbf{w/o u}: Without using the unified data preprocessing approach (detailed in~\ref{adj}).
\end{itemize}

The results are shown in Figure~\ref{ablation}.
We summarize our findings as follows:
(1) Geo-STARS outperforms all the ablated variants across both settings, confirming the effectiveness of each component in Geo-STARS and demonstrating its strength in simultaneously capturing spatial-temporal dynamics by leveraging the geographic consistency. (2) The removal of geo-aware embedding (w/o z) leads to 
the worst 
zero-shot performance, underscoring the critical role of the geo-aware embedding in enabling generalization its deep integration into the model. (3) The degraded performance achieved by w/o c demonstrates the function of original stream characteristics in calculating gate weights. (4) Both variants by removing adjacency matrices (w/o A' and w/o A), show performance deterioration, especially in fine-tuning. This demonstrates that it is essential to consider the information from both the previous day and the current day. (5) The variant w/o u performs worse than Geo-STARS, highlighting the significance of properly processing the distance matrix.

\subsection{Missing Stream Characteristics}

\begin{figure}[t!]
\centering
\includegraphics[width=\columnwidth]
{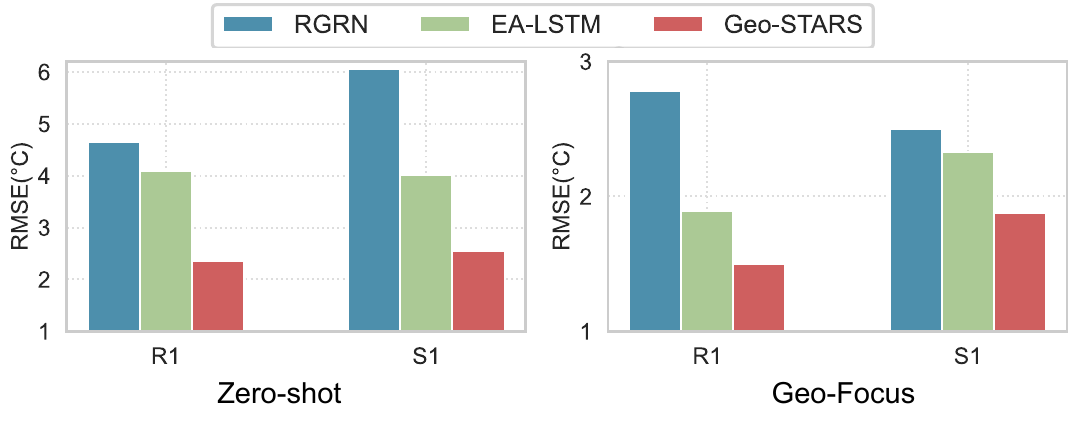}
\caption{Comparison of average RMSE for stream water temperature prediction under the missing characteristics scenario. Left: zero-shot results for R1 and S1. Right: geo-aware embedding fine-tuning results for R1 and S1.}
\label{missingC}
\end{figure}

Due to the technical difficulties or the high cost of measuring certain stream characteristics, the full set of characteristics may not be available across all the watersheds. Therefore, we consider a scenario in which the target tasks have incomplete characteristics. 
To simulate this, we randomly mask out 55 out of 65 available characteristic features 
 (approximately 85\%) for each target dataset. This masking is performed independently for each dataset to mimic diverse missing patterns.
We select experiments R1 and S1 for this test to evaluate whether our model can generalize across both regions and scales. Fig.~\ref{missingC} presents the results of Geo-STARS alongside the other two best-performing baselines: RGRN and EA-LSTM. Geo-STARS consistently outperforms these baselines when using either no observations (zero-shot) or small observations 
(Geo-Focus)
for tuning the model. This shows that Geo-STARS can be well generalized to other tasks even when extensive stream characteristics are unavailable for the target task. This advantage originates from the ability of Geo-STARS to effectively utilize stream characteristics. Most baseline methods directly use characteristics as inputs, leading to performance degradation when variables are missing and set to default values. In contrast, Geo-STARS uses characteristics in gating and geo-aware embedding calculations, avoiding this issue. As a result, the absence of some variables has a less negative impact. Furthermore, Geo-Focus in Geo-STARS can effectively use limited observations to compensate for the missing characteristics.

\section{Conclusion}
In this study, we introduced Geo-STARS, a novel geo-aware framework designed to generalize stream temperature predictions across different regions and spatial scales, even under sparse or zero observation settings. By integrating geo-aware embeddings with a gated spatio-temporal graph neural network, Geo-STARS effectively captures both shared geographic principles and spatial-temporal patterns specific to each learning task. 
In experiments, Geo-STARS is shown to outperform existing methods in predicting water temperatures across diverse watersheds and different spatial scales. The superior predictive performance highlights the potential of geographic information in improving generalization ability, which further offers advantages for more informed decision-making in resource management and ecosystem preservation for large regions at high spatial resolutions.

Future work will focus on further exploring the potential of the proposed method as a foundational model for environmental ecosystems by incorporating additional geographic features and extending the model to predict a wide range of relevant variables in the target environmental ecosystem.  A promising direction includes extending the model to predict multiple water quality variables (e.g., water temperature, dissolved oxygen) and water quantity variables (e.g., streamflow), which could offer a more comprehensive perspective to support water management efforts.

\section{Acknowledgments}
This work was supported by the National Science Foundation (NSF) under grants 2239175, 2316305, 2147195, 2425844, 2425845, 2430978, 2126474, and 2530610;
the USGS awards  G21AC10564 and G22AC00266;   the NASA grants 80NSSC24K1061 and 80NSSC25K0013; and the NSF NCAR's Derecho HPC system. This research was also supported in part by the University of Pittsburgh Center for Research Computing through the resources provided.



\bibliographystyle{ACM-Reference-Format}
\bibliography{reference, Xiaowei, shengyu}


\clearpage
\appendix

\balance

\section{Main Results}
Due to space constraints, we only present the average results of the target tasks in each experiment within the main content. Here, Table~\ref{result_sparsity_1} and Table~\ref{result_sparsity_2} provide the detailed results for each target task across all experiments at a sparsity level of 0.1\%, corresponding to Table 2 in the main content, which shows the average RMSE of the target tasks(e.g., the results of R1 are the average value of UL$_\text{c}$ and UL$_\text{f}$). Similarly, Table~\ref{result_zero_1} and Table~\ref{result_zero_2} show the detailed zero-shot RMSE results, corresponding to Figure 4 in the main content, which is drawn based on the average RMSE values of the target tasks.

\begin{table*}[!b]
\centering
\small
\caption{Comparison of average RMSE for stream water temperature prediction under the sparsity level of 0.1\% for experiments in R1 to R4. The best results are bold.}
\begin{tabular}{llccccccccccc}
\toprule
    & & \multicolumn{2}{c|}{\textbf{R1}} & \multicolumn{2}{c|}{\textbf{R2}} & \multicolumn{2}{c|}{\textbf{R3}} & \multicolumn{2}{c}{\textbf{R4}} \\
    \cmidrule(lr){3-4} \cmidrule(lr){5-6} \cmidrule(lr){7-8} \cmidrule(lr){9-10} 
    \textbf{Methods} & \textbf{Strategies} & UL$_\text{c}$ & UL$_\text{f}$ & NS$_\text{c}$ & NS$_\text{f}$ & LV$_\text{c}$ & LV$_\text{f}$ & CRW$_\text{c}$ & CRW$_\text{f}$ \\
    \midrule
    LSTM & Complete & 3.55 & 2.53 & 3.04 & 2.70 & 2.68 & 3.70 & 2.84 & 3.34 \\
    Transformer & Complete & 2.97 & 3.88 & 2.81 & 2.70 & 3.39 & 3.45 & 2.87 & 2.68 \\
    RGRN & Complete & 2.27 & 1.97 & 2.33 & 2.63 & 3.04 & 3.94 & 3.28 & 3.05 \\
    GCN-LSTM & Complete & 2.91 & 3.15 & 3.10 & 3.13 & 3.69 & 4.20 & 3.79 & 3.75 \\
    Mini-Batch & Complete & 2.13 & 2.08 & 2.55 & 2.95 & 3.05 & 3.73 & 3.57 & 2.93 \\
    Graph-WaveNet & Complete & 2.74 & 2.73 & 3.19 & 3.51 & 3.44 & 4.30 & 3.65 & 3.55 \\
    EA-LSTM & Complete & 1.90 & 2.08 & 2.20 & 2.34 & 3.14 & 3.31 & 2.62 & 2.81 \\
    SR-MTL & Complete & 2.14 & 2.15 & 2.50 & 2.76 & 2.99 & 3.41 & 2.03 & 2.15 \\
    STCGAN & Complete & 2.01 & 2.23 & 2.03 & 2.11 & 2.98 & 3.31 & 2.80 & 2.37 \\
    \cmidrule{1-10}
    \multirow{3}{*}{Geo-STARS} & Complete & \textbf{1.69} & \textbf{1.45} & 2.11 & \textbf{1.85} & \textbf{2.68} & 2.92 & \textbf{2.19} & 1.97 \\
    & Geo-Related & 1.90 & 1.69 & 2.19 & 2.05 & 2.79 & \textbf{2.80} & 2.57 & \textbf{1.89} \\
    & Geo-Focus & 1.96 & 1.78 & \textbf{1.89} & 2.16 & 3.49 & 4.27 & 2.61 & 2.91 \\
    \bottomrule
\end{tabular}
\label{result_sparsity_1}
\end{table*}

\begin{table*}[!b]
\centering
\small
\caption{Comparison of average RMSE for stream water temperature prediction under the sparsity level of 0.1\% for experiments in S1 to S4. The best results are bold.}
\begin{tabular}{llcccccccccccc}
    \toprule
    & & \multicolumn{2}{c|}{\textbf{S1}} & \multicolumn{2}{c|}{\textbf{S2}} & \multicolumn{3}{c|}{\textbf{S3}} & \multicolumn{3}{c}{\textbf{S4}} \\
    \cmidrule(lr){3-4} \cmidrule(lr){5-6} \cmidrule(lr){7-9} \cmidrule(lr){10-12} 
    \textbf{Methods} & \textbf{Strategies} & NS$_\text{f}$ & UL$_\text{f}$ & CRW$_\text{f}$ & LV$_\text{f}$ & NS$_\text{f}$ & CRW$_\text{f}$ & UL$_\text{f}$ & CRW$_\text{f}$ & UL$_\text{f}$ & LV$_\text{f}$ \\
    \midrule
    LSTM & Complete & 2.42 & 3.28 & 2.40 & 4.10 & 2.75 & 2.51 & 3.76 & 2.21 & 2.30 & 4.41 \\
    Transformer & Complete & 3.04 & 3.07 & 2.55 & 3.53 & 2.90 & 2.51 & 2.82 & 2.80 & 3.06 & 3.62 \\
    RGRN & Complete & 2.24 & 2.42 & 3.25 & 3.27 & 3.07 & 3.15 & 2.42 & 3.29 & 3.10 & 4.07 \\
    GCN-LSTM & Complete & 3.65 & 3.16 & 3.31 & 3.77 & 3.18 & 3.96 & 3.31 & 4.26 & 3.68 & 4.13 \\
    Mini-Batch & Complete & 2.94 & 2.11 & 3.52 & 4.15 & 2.66 & 2.82 & 2.05 & 4.18 & 2.75 & 3.52 \\
    Graph-WaveNet & Complete & 2.99 & 3.10 & 2.89 & 3.80 & 2.20 & 3.83 & 2.90 & 3.70 & 3.20 & 4.03 \\
    EA-LSTM & Complete & 2.52 & 2.21 & 2.81 & 3.50 & 2.61 & 2.98 & 2.25 & 2.93 & 2.28 & 3.00 \\
    SR-MTL & Complete & 2.66 & 2.83 & 2.34 & 3.48 & 2.31 & 2.36 & 2.42 & 2.68 & 2.40 & 3.45 \\
    STCGAN & Complete & 3.57 & 3.41 & 2.70 & 3.37 & 2.19 & 2.66 & 2.85 & 3.52 & 2.96 & 3.67 \\
    \cmidrule{1-12}
    \multirow{3}{*}{Geo-STARS} & Complete & \textbf{1.73} & 1.53 & \textbf{1.92} & \textbf{3.25} & 1.76 & \textbf{1.88} & 1.56 & \textbf{1.88} & 1.64 & \textbf{2.98} \\
    & Geo-Related & 1.82 & \textbf{1.45} & 2.20 & 3.37 & 2.08 & 2.15 & \textbf{1.63} & 2.05 & 1.90 & 3.01 \\
    & Geo-Focus & 1.75 & 1.44 & 2.48 & 3.77 & \textbf{1.93} & 2.26 & 1.64 & 2.17 & \textbf{1.62} & 3.46 \\
    \bottomrule
\end{tabular}
\label{result_sparsity_2}
\end{table*}

\begin{table*}[!t]
\centering
\small
\caption{Average RMSE of zero-shot setting for stream water temperature prediction for experiments in R1 to R4. The best results are bold.}
\begin{tabular}{lc|c|c|c|c|c|c|c}
\toprule
& \multicolumn{2}{c|}{\textbf{R1}} & \multicolumn{2}{c|}{\textbf{R2}} & \multicolumn{2}{c|}{\textbf{R3}} & \multicolumn{2}{c}{\textbf{R4}} \\
\cmidrule(lr){2-3} \cmidrule(lr){4-5} \cmidrule(lr){6-7} \cmidrule(lr){8-9} 
& UL$_\text{c}$ & UL$_\text{f}$ & NS$_\text{c}$ & NS$_\text{f}$ & LV$_\text{c}$ & LV$_\text{f}$ & CRW$_\text{c}$ & CRW$_\text{f}$ \\
\midrule
LSTM & 4.07 & 3.32 & 3.36 & 4.87 & 4.32 & 4.79 & 6.16 & 5.85 \\
Transformer & 8.01 & 4.22 & 3.78 & 4.22 & 4.69 & 6.15 & 7.35 & 5.12 \\
RGRN & 3.29 & 3.97 & 4.10 & 5.08 & 5.04 & 7.10 & 5.36 & 5.51 \\
GCN-LSTM & 3.79 & 4.26 & 4.49 & 6.84 & 7.93 & 7.93 & 5.67 & 6.82 \\
Mini-Batch & 6.51 & 6.35 & 4.06 & 4.67 & 5.71 & 7.16 & 5.40 & 6.29 \\
Graph-WaveNet & 6.88 & 6.53 & 4.79 & 5.81 & 5.33 & 6.63 & 6.26 & 6.42 \\
EA-LSTM & 5.37 & 3.88 & 3.37 & 3.82 & 4.17 & 5.13 & 4.86 & 4.71 \\
SR-MTL & 4.16 & 3.07 & 4.27 & 5.68 & \textbf{4.16} & 4.66 & 4.37 & 4.47 \\
STCGAN & 2.36 & 3.86 & 3.09 & 3.46 & 4.77 & 5.92 & 4.36 & 5.53 \\
\cmidrule{1-9}
Geo-STARS & \textbf{2.92} & \textbf{2.26} & \textbf{2.61} & \textbf{2.91} & 4.31 & \textbf{4.50} & \textbf{3.46} & \textbf{4.44} \\
\bottomrule
\end{tabular}
\label{result_zero_1}
\end{table*}

\begin{table*} [!t]
\centering
\small
\caption{Average RMSE of zero-shot setting for stream water temperature prediction for experiments in S1 to S4. The best results are bold.}
\begin{tabular}{lc|c|c|c|c|c|c|c|c|c}
\toprule
& \multicolumn{2}{c|}{\textbf{S1}} & \multicolumn{2}{c|}{\textbf{S2}} & \multicolumn{3}{c|}{\textbf{S3}} & \multicolumn{3}{c}{\textbf{S4}} \\
\cmidrule(lr){2-3} \cmidrule(lr){4-5} \cmidrule(lr){6-8} \cmidrule(lr){9-11}
& NS$_\text{f}$ & UL$_\text{f}$  & CRW$_\text{f}$ & LV$_\text{f}$  & NS$_\text{f}$ & CRW$_\text{f}$ & UL$_\text{f}$  & CRW$_\text{f}$ & UL$_\text{f}$ & LV$_\text{f}$  \\
\midrule
LSTM & 3.77 & 4.10 & 5.18 & 5.08 & 3.34 & 5.39 & 4.29 & 5.36 & 5.77 & 3.82 \\
Transformer & 5.03 & 4.65 & 5.14 & 5.51 & 6.19 & 4.49 & 4.14 & 5.4 & 4.99 & 6.67 \\
RGRN & 3.95 & 4.81 & 4.56 & 5.94 & 3.27 & 3.98 & 4.10 & 4.17 & 3.86 & 5.04 \\
GCN-LSTM & 6.48 & 4.82 & 6.72 & 6.97 & 6.04 & 5.99 & 6.14 & 6.75 & 7.05 & 6.26 \\
Mini-Batch & 4.50 & 3.81 & 4.94 & 5.64 & 3.99 & 3.76 & 3.84 & 5.64 & 6.22 & 5.08 \\
Graph-WaveNet & 5.83 & 3.85 & 5.53 & 5.01 & 5.88 & 6.79 & 3.51 & 5.36 & 7.49 & 5.35\\
EA-LSTM & 3.26 & 3.56 & 3.60 & 5.32 & 3.80 & 4.44 & 4.16 & 4.29 & 3.68 & 3.56 \\
SR-MTL & 4.80 & 6.84 & 4.88 & 4.82 & 5.39 & 3.78 & 3.19 & 5.47 & 3.19 & 4.28 \\
STCGAN & 4.16 & 3.76 & 3.69 & 4.38 & 3.79 & 3.66 & 2.71 & 3.94 & 4.74 & 4.54 \\
\cmidrule{1-11}
Geo-STARS & \textbf{1.90} & \textbf{1.44} & \textbf{2.90} & \textbf{4.05} & \textbf{2.27} & \textbf{2.77} & \textbf{1.86} & \textbf{2.39} & \textbf{2.21} & \textbf{3.45} \\
\bottomrule
\end{tabular}
\label{result_zero_2}
\end{table*}

\end{document}